\documentclass[letterpaper]{article} 
\usepackage[draft]{aaai2026}  

\usepackage{times}  
\usepackage{helvet}  
\usepackage{courier}  
\usepackage[hyphens]{url}  
\usepackage{graphicx} 
\usepackage{natbib}  
\usepackage{caption} 
\usepackage{algorithm}
\usepackage{algorithmic}
\usepackage{xcolor}
\usepackage{newfloat}
\usepackage{listings}
\DeclareCaptionStyle{ruled}{labelfont=normalfont,labelsep=colon,strut=off} 
\floatstyle{ruled}
\newfloat{listing}{tb}{lst}{}
\floatname{listing}{Listing}
\title{LEAML: Label-Efficient Adaptation to Out-of-Distribution \\ Visual Tasks for Multimodal Large Language Models}
\author{Ci-Siang Lin$^{1}$ \qquad Min-Hung Chen$^{2}$ \qquad Yu-Yang Sheng$^{1}$ \qquad Yu-Chiang Frank Wang$^{1, 2}$\\
$^{1}$Graduate Institute of Communication Engineering, National Taiwan University, Taiwan \qquad $^{2}$NVIDIA\\
}
\usepackage{bibentry}
\usepackage{amsmath}
\usepackage{amsfonts}
\usepackage{multirow}
\usepackage{booktabs}
\newcommand{\nameshort}{\textcolor{black}{LEAML}}
\newcommand{\namestagea}{\textcolor{black}{Pseudo QA Generation}}
\newcommand{\namestageaa}{\textcolor{black}{Selective Neuron Distillation}}
\newcommand{\namestageb}{\textcolor{black}{OOD VQA Finetuning}}

\begin{document}

\maketitle
\begin{abstract}
Multimodal Large Language Models (MLLMs) have achieved strong performance on general visual benchmarks but struggle with out-of-distribution (OOD) tasks in specialized domains such as medical imaging, where labeled data is limited and expensive. We introduce \nameshort, a label-efficient adaptation framework that leverages both scarce labeled VQA samples and abundant unlabeled images. Our approach generates domain-relevant pseudo question-answer pairs for unlabeled data using a QA generator regularized by caption distillation. Importantly, we selectively update only those neurons most relevant to question-answering, enabling the QA Generator to efficiently acquire domain-specific knowledge during distillation. Experiments on gastrointestinal endoscopy and sports VQA demonstrate that \nameshort~consistently outperforms standard fine-tuning under minimal supervision, highlighting the effectiveness of our proposed \nameshort~framework.
\end{abstract}
\section{Introduction}
Large Language Models (LLMs)~\cite{touvron2023llama, achiam2023gpt, bai2023qwen} have demonstrated impressive capabilities across diverse language tasks. By incorporating visual understanding, Multimodal Large Language Models (MLLMs)~\cite{alayrac2022flamingo, liu2023visual,wang2024qwen2} extend these capabilities to visual question answering (VQA), image captioning, and multimodal reasoning tasks.~\cite{wang2024exploring} Recent MLLMs have achieved remarkable performance on general visual benchmarks, showing strong generalization within their training distributions. Although scaling laws~\cite{kaplan2020scaling} show promise for enhancing model performance, they do not eliminate dependence on pre-training data, which limits generalization to novel domains. In real-world deployment, MLLMs will face domain-specific tasks that fall outside their training distribution, such as specialized medical imaging or technical visual content. When confronted with such out-of-distribution (OOD) data, these models often produce erroneous or unreliable outputs, limiting their applicability in specific domains where accuracy is critical.

While existing studies on OOD have primarily focused on detection~\cite{ming2022delving, wang2023clipn} or domain-specific classification tasks~\cite{liu2023clip, lei2023clip, zhang2023biomedclip}, applying MLLMs to visual question answering in unfamiliar domains introduces distinct challenges. Unlike detection, which focuses on filtering unfamiliar inputs, VQA requires models to generate accurate and contextually grounded answers for images outside the training distribution. Although many specific domains offer abundant unlabeled images, standard VQA training pipelines depend on paired question-answer annotations, which are costly to obtain due to the need for domain expertise. Furthermore, VQA questions must be carefully constructed based on the visual content. Randomly pairing unrelated questions with images creates confusing training signals and reduces the effectiveness of learning. Due to the scarcity of high-quality labeled data, fully fine-tuning MLLMs often leads to severe overfitting, particularly in VQA where models must produce free-form, semantically appropriate responses rather than select from fixed labels. These challenges underscore the need for adaptation methods that can leverage both limited labeled examples and large pools of unlabeled images in a way that preserves generation capabilities while acquiring domain-specific understanding.

In this paper, we introduce \textbf{\nameshort}, a two-stage \textbf{L}abel-\textbf{E}fficient \textbf{A}daptation framework for \textbf{M}ultiModal \textbf{L}LM designed to effectively leverage both the limited labeled VQA data and the abundant unlabeled images in the target domain. The framework comprises \namestagea, which constructs domain-relevant QA pairs from unlabeled images, and \namestageb, which fine-tunes the target MLLM using both the generated pairs and the original labeled data. In \namestagea, a QA Generator is trained using the small labeled dataset to capture the domain-specific patterns of question-answer formulations. To mitigate overfitting caused by the scarcity of labeled data, the generator is regularized via caption distillation, where it additionally learns from image captions produced by a large-scale model, providing broader visual-linguistic signals beyond the limited annotations. The generator then synthesizes diverse pseudo QA pairs for the unlabeled domain images, creating a significantly richer training resource. In \namestageb, the target MLLM is fine-tuned with both the original labeled samples and the generated QA pairs. To enhance adaptation efficiency and prevent overfitting, we employ \namestageaa, which is motivated by the insight that domain-specific knowledge is often encoded in a subset of neurons; focusing updates on these neurons enables the model to acquire domain-relevant reasoning capabilities while preserving general language generation skills. Experiments conducted on gastrointestinal endoscopy VQA demonstrate that our method significantly improves performance compared to standard fine-tuning approaches.

Our contributions can be summarized as follows:
\begin{itemize}
    \item We propose a two-stage learning framework \nameshort, which leverages both scarce labeled data and abundant unlabeled images to achieve MLLM adapataion for domain-specific VQA .
    \item We introduce \namestagea, which learns a generator to produce pseudo question-answer pairs for unlabeled data, augmenting the training set for finetuning the VQA model on specialized domains.
    \item We design \namestageaa~for the QA generator, which performs captioning distillation to acquire domain-related knowledge while selectively updates QA-related neurons, resulting in reliable pseudo QA pairs for finetuning.

\end{itemize}

\section{Related Works}
\subsection{Out-of-Distribution Data Learning}
Out-of-distribution (OOD) data refers to inputs that differ significantly from those seen during training. Early OOD research~\cite{ming2022delving, wang2023clipn} primarily focused on detection tasks, where the goal is to identify whether an input falls outside the training distribution. These methods often treat OOD detection as anomaly detection. However, these methods aim to identify and reject OOD inputs rather than adapt models to perform well on them. 
Beyond detection, another line of work explores how vision-language models can be adapted for OOD domains. CLIP and its variants have shown promise in medical imaging applications through domain-specific fine-tuning~\cite{liu2023clip,lei2023clip, zhang2023biomedclip}. 
However, these approaches are limited to classification settings, where models predict labels from a fixed set rather than generating open-ended responses.

Recent works explore adapting LLMs to specific domains~\cite{zhang2023reformulating, cheng2024domain, bhatia2024fintral,cheng2023adapting}. For instance,~\cite{zhang2023reformulating, kim2025videoicl} employ retrieval-based approaches to handle out-of-distribution inputs, relying on external sources to provide relevant context for domain adaptation. However, such methods still assume that the LLM has sufficient foundational knowledge to effectively interpret the retrieved information.
In the vision-language domain, ~\cite{cheng2024domain} propose fine-tuning MLLMs to automatically generate question-answer pairs from existing image-caption datasets, subsequently using these synthetic QA pairs for domain-specific VQA training. However, the effectiveness of such methods depends heavily on both the LLM's domain knowledge and the granularity of source captions. Fine-grained captions enable detailed QA generation, while coarse-grained descriptions yield only generic questions. These factors highlight the need for methods that can produce high-quality domain-specific QA pairs even with limited domain knowledge and varying caption quality.

\subsection{Semi-Supervised Learning}
Semi-supervised learning (SSL) is a strategy to bridge the gap between limited labeled data and abundant unlabeled resources, especially in domains where annotation is expensive or requires domain expertise. Among existing SSL paradigms, pseudo-labeling~\cite{lee2013pseudo,xie2020self} has been widely adopted. This approach trains models on labeled data, then uses them to pseudo-label unlabeled samples for further training.

Pseudo-labeling has been widely applied to tasks such as image classification~\cite{zeng2023pefat} and Segmentation~\cite{yang2023revisiting}, where labels are either discrete or can be directly generated from the data. In these settings, unlabeled samples can be automatically annotated with relatively reliable supervision. In contrast, applying pseudo-labeling to visual question answering (VQA) is considerably more difficult. VQA requires not only a grounded answer but also a relevant and context-aware question that must be closely aligned with the visual content. Unlike captions or class labels, such question-answer pairs cannot be directly inferred from the image alone. A naïve solution might involve randomly assigning questions to domain images, but this leads to incoherent or misleading training data. As a result, semi-supervised learning remains largely underexplored in VQA, particularly in domain-specific scenarios where labeled data is scarce and generating valid pseudo QA pairs is highly non-trivial.


\subsection{Neuron-Level Knowledge Attribution in DNNs}
Knowledge in neural networks is localized and stored within specific neural components rather than distributed uniformly across parameters. 
Previous work has shown that the feed-forward network (FFN) layers in Transformers play a key role in storing knowledge~\cite{geva2020transformer,geva2022transformer}. These layers have been characterized as performing additive, knowledge-based updates on token representations. \cite{dai2021knowledge} propose a neuron-level attribution approach to identify knowledge neurons in FFNs responsible for storing factual knowledge.
These findings indicate that MLLMs encode different types of knowledge in specialized neural components with distinct activation patterns, enabling targeted manipulation of specific knowledge without affecting the entire network.

\begin{figure*}[!t]
  \centering
  \includegraphics[width=\linewidth]{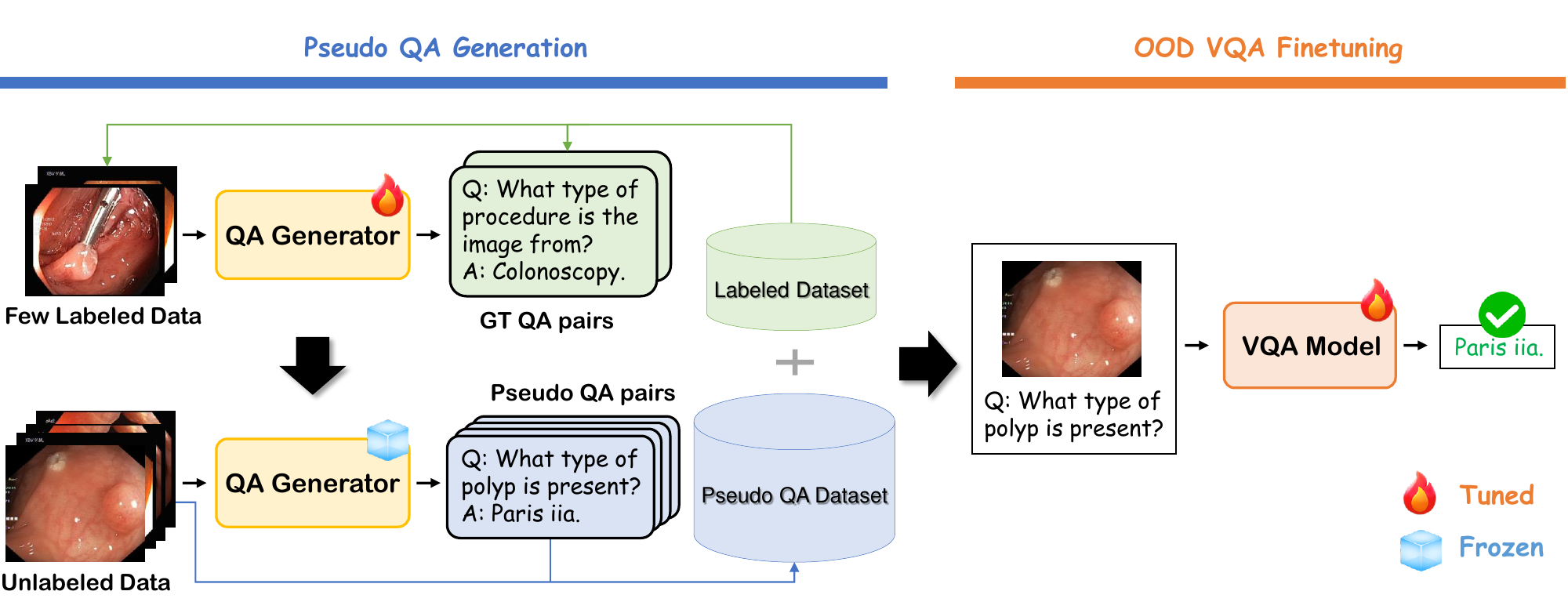}
  \caption{Overview of the proposed two-stage \nameshort~framework for OOD VQA adaptation. In \namestagea, the QA Generator is trained using a small set of labeled question-answer pairs and then used to generate pseudo QA pairs for a large collection of unlabeled images. In \namestageb, the VQA model is fine-tuned with both the original labeled data and the produced pseudo QA pairs of unlabeled data, enabling label-efficient adaptation to out-of-distribution visual-question answering. We will detail the learning of our QA Generator in Figure~\ref{figure:model2}.}
  \label{figure:model1}
\end{figure*}
To support such targeted interventions, a variety of attribution methods have been proposed to quantify the importance of individual neurons or weights. Some are designed to guide pruning, which reduces model size by removing components deemed less relevant. For example, Wanda~\cite{sun2023simple} identifies unimportant weights by combining their magnitude with the norm of the associated input activation. Others~\cite{fang2024towards,pan2023finding,yu2023neuron,yu2025locate} aim to inform fine-tuning, where only selected parts of the model are updated to acquire new capabilities or adapt to specific domains. A common strategy in this context is to measure gradient magnitudes during backpropagation, under the assumption that neurons with higher gradients contribute more significantly to the output~\cite{zhang2024gradient}.


\section{Method}

\subsection{Problem Formulation and Framework Overview}

\paragraph{Problem Formulation.} We first define the problem settings and notations used in this paper. For the out-of-distribution visual question-answering problem, we consider domain-specific tasks (e.g., gastrointestinal endoscopy) which are still based on RGB images or videos but are rarely or not covered in the pretraining data of general-purpose multimodal large language models. Since such domain-specific tasks generally require great expense to obtain manual annotations from domain experts, we further consider a realistic yet challenging setting, where only few data are annotated in the training dataset $D$. That is, the training dataset $D$ contains a set of labeled data $D_l$ of few samples and also a set of abundant unlabeled data $D_u$, which is similar to traditional semi-supervised learning. Each data instance in the labeled dataset $D_l$ contains an image or video $V$, an associated question $Q$, and the corresponding answer $A$, while each sample in the unlabeled dataset $D_u$ contains only the visual part $V$.

\paragraph{Framework Overview.} Given the above training data and a pretrained general MLLM, our goal is to adapt this pretrained MLLM to address out-of-distribution visual question-answering with only few labeled instances. To achieve this goal, we propose a two-stage learning framework \nameshort~as shown in Figure~\ref{figure:model1}. Our \nameshort~framework includes two stages: \namestagea~and \namestageb. The former stage aims to generate proper pseudo question-answer pairs for the unlabeled dataset $D_u$, while the latter stage leverages both the labeled dataset $D_l$ and the produced pseudo QA pairs for the unlabeled data $D_u$ to finetune a pretrained MLLM for addressing out-of-distribution visual question-answering. 

\begin{figure*}[!t]
  \centering
  \includegraphics[width=\linewidth]{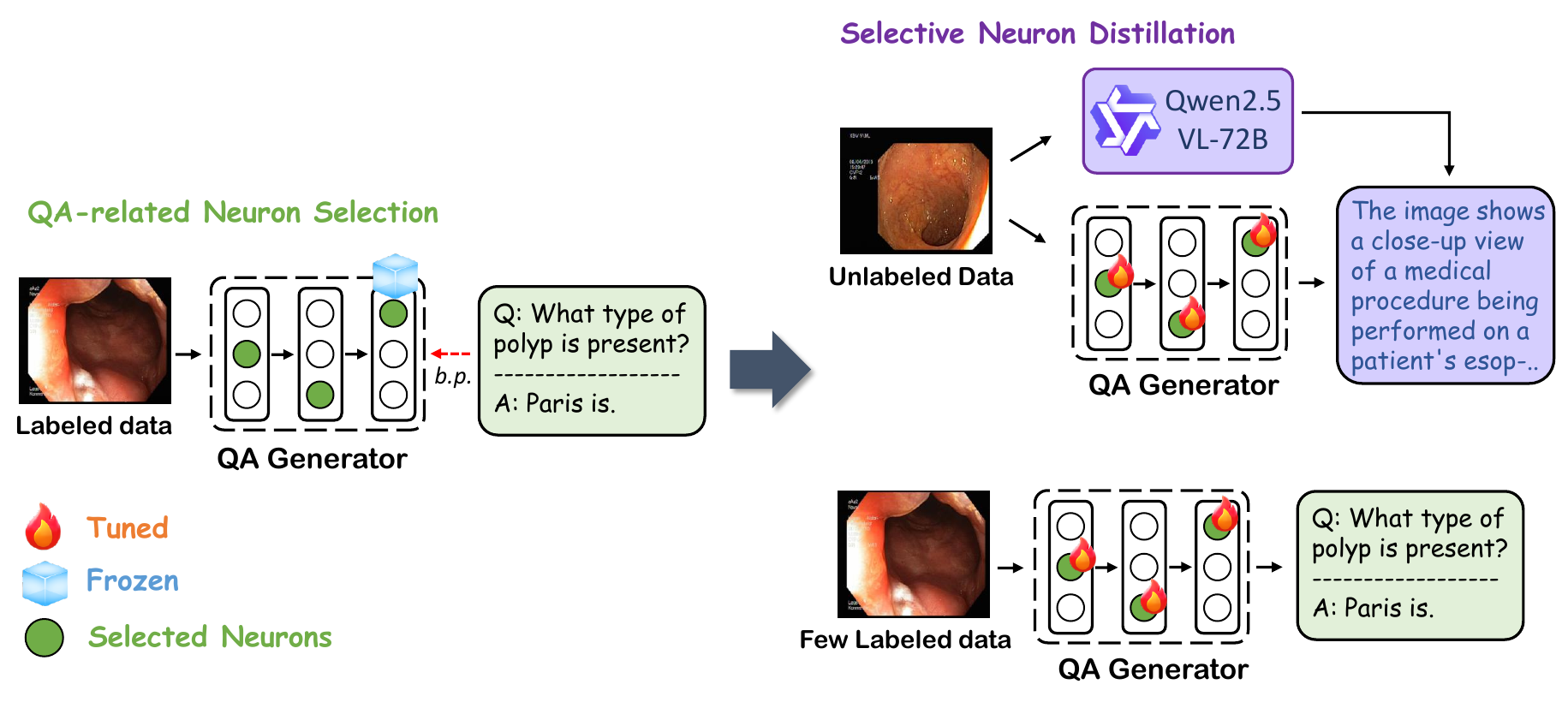}
  \caption{Illustration of our \namestageaa~for the QA Generator. The QA-relevant parameters are first selected based on gradient scores from labeled QA data. During training, only these selected parameters are updated using auxiliary caption supervision from unlabeled images, allowing QA-related knowledge distillation for the QA Generator.}
  \label{figure:model2}
\end{figure*}

Specifically, since the unlabeled data only contain the visual information but lack the corresponding textual annotations for training a VQA model, our \nameshort~framework first employ a QA Generator $G$ which is directly supervised by the labeled data to produce pseudo QA pairs for unlabeled data. However, in this manner, the QA Generator may overfit on those few labeled samples and fail to generalize well to produce reliable QA pairs for unlabeled data. To overcome this challenge, we further propose \namestageaa~to leverage also the unlabeled data during the training of the QA Generator. As shown in Figure~\ref{figure:model2}, in addition to the QA generation supervised by labeled data, we perform captioning distillation using unlabeled data to distill related knowledge from a large MLLM. More importantly, during training, we update only QA-related neurons for both the QA generation and captioning distillation. In this way, such captioning distillation is designed to focus solely on gaining QA-related knowledge, and hence the QA Generator $G$ is able to produce reliable pseudo QA pairs for unlabeled data during inference, benefiting the following finetuning for the VQA model. We now detail our learning framework in the following subsections.

\subsection{\namestagea~and \namestageb}

Recently, multimodal large language models (MLLMs) have achieved impressive generalization on a wide range of vision-language tasks. However, their performance often degrades when applied to specialized domains such as medical imaging, where domain-specific visual cues and terminology differ significantly from pretraining data. A significant barrier in adapting multimodal large language models (MLLMs) to these specialized domains is the extreme scarcity of annotated question-answer (QA) pairs. Obtaining such labeled data from domain experts is often expensive, time-consuming, and infeasible at scale, leading to severe limitations on direct fine-tuning. On the other hand, large volumes of unlabeled visual data typically exist, representing a potential yet largely untapped source of supervision. While traditional semi-supervised learning have demonstrated that unlabeled data can be highly beneficial, such success is largely limited to classification tasks. For unlabeled data in visual question-answering, both the question and answer are not accessible, and hence traditional semi-supervised learning cannot be easily applied in VQA to generate the pseudo labels without knowing or providing a corresponding question.


To overcome the limitation of requiring fully annotated data, we introduce a pseudo QA generation stage that transforms unlabeled visual data into useful supervision for the VQA model. The core idea is to train a QA Generator using limited labeled examples $D_l$, and then apply it to unlabeled inputs $D_u$ to produce additional VQA samples. Specifically, for each labeled instance $(V, Q, A)$ in $D_l$, we formulate the corresponding target text sequence $Y$ by concatenating of the question $Q$ and answer $A$ with special tokens: 

\begin{equation}
\begin{aligned}
Y = [\texttt{<q>} Q \texttt{<q>} \texttt{<a>} A \texttt{<a>}].
\end{aligned}
\end{equation}
Then, the QA generator $G$, initialized from a pretrained general MLLM, is trained to autoregressively generate $Y$ conditioned on the visual input $V$. The training objective $\mathcal{L}_{QA}$ is defined as a standard autoregressive negative log-likelihood:

\begin{equation}
\begin{aligned}
\mathcal{L}_{QA} = \sum_{t=1}^{|Y|} \log p(y_i \mid V, y_{<i}), V \in D_l
\end{aligned}
\label{eq:qa}
\end{equation}
where $y_i$ denotes the $i$th text token of the sequence $Y$ and $p$ is the likelihood of text tokens estimated by the QA Generator $G$. Once the training is complete, the QA generator $G$ would be applied to produce pseudo QA pairs $[\texttt{<q>} \hat{Q} \texttt{<q>} \texttt{<a>} \hat{A} \texttt{<a>}] = G(V)$ for the visual input $V$ from unlabeled data, resulting in the dataset $\hat{D_{u}} = \{(V, \hat{Q}, \hat{A})\}$. This dataset is then used to augment the original labeled data $D_l$ for finetuning the final VQA model with similar autoregressive objective $\mathcal{L}_{VQA}$:

\begin{equation}
\begin{aligned}
\mathcal{L}_{VQA} &= \sum_{t=1}^{|A|} \log p(a_i \mid V, Q, a_{<i}), V \in D_l, \hat{D}_u
\end{aligned}
\label{eq:vqa}
\end{equation}
Note that the VQA model is again initialized from a pretrained general MLLM. While the above pseudo QA generation enables the model to leverage unlabeled visual data for OOD VQA adaptation, the quality of generated QA pairs is tightly bound to the small number of labeled examples. With limited supervision, the model may produce noisy outputs on unlabeled data, especially under the out-of-distribution settings.

\subsection{\namestageaa~for QA Generator}
To further mitigate the aforementioned problem of \namestagea, our \nameshort~framework introduces a distillation mechanism that directly incorporates abundant unlabeled data into the QA Generator training process. In addition to the QA generation part supervised by labeled data $D_l$, we further consider captioning distillation on unlabeled $D_u$ to gain domain-related knowledge from a large MLLM. Specifically, for each unlabeled visual input $V$, we employ a large pre-trained general MLLM to automatically generate a descriptive caption $C$. While the large MLLM may still be poor when directly applied to perform OOD VQA tasks, these state-of-the-art large models are pre-trained on vast and diverse corpora such as medical literature, Wikipedia articles, and other authoritative web resources. As a result, the captions generated by such models are often infused with rich semantic information, domain-specific terminology, and contextual knowledge, which are beneficial to the learning of the QA Generator $G$. Formally, the QA Generator $G$ is jointly optimized with the QA objective $\mathcal{L}_{QA}$ in Equation~\ref{eq:qa} using labeled data $D_l$ and the captioning distillation objective $\mathcal{L}_{C}$ using unlabeled data $D_u$:

\begin{equation}
\begin{aligned}
\mathcal{L}_{C} = - \sum_{t=1}^{|C|} &\log p(c_i \mid V, c_{<i}), V \in D_u \\
\mathcal{L}_{G} &= \mathcal{L}_{QA} + \mathcal{L}_{C}
\end{aligned}
\end{equation}

While caption distillation enables the QA Generator $G$ to absorb rich semantic knowledge from unlabeled data, it remains essential that the model's capacity is focused on generating question-answer pairs rather than captioning. Under this motivation, we introduce a neuron selection strategy to ensure such captioning distillation is solely for enhancing QA-related knowledge. Specifically, we identify parameters in the QA Generator that contribute most to the question-answer generation and restrict parameter updates during training to this subset. For each parameter $\theta$ in the QA Generator $G$, its importance score $s$ is quantified by the average magnitude of the gradient of the QA loss $\mathcal{L}_{QA}$:

\begin{equation}
s = \left| \frac{1}{|D_l|} \sum_{(V, Q, A) \in D_l} \frac{\partial \mathcal{L}_{QA}}{\partial \theta} \right|.
\end{equation}
Then, we selection and update only parameters with top-$K$ scores in each neuron (i.e., each row in a linear weight matrix) while keep others frozen:

\begin{equation}
\theta \leftarrow
\begin{cases}
\theta - \eta \cdot \frac{\partial \mathcal{L}_{G}}{\partial \theta} & \text{if $s$ is among top-$K$}  \\
\theta, & \text{otherwise}
\end{cases}
\end{equation}
By restricting updates to only the most QA-relevant neurons or parameters, we ensure that the auxiliary knowledge gained from caption distillation is efficiently integrated to support question-answer generation, rather than generic captioning or unrelated model capacities. With the above learning, the QA Generator is not limited to patterns seen in the small labeled set, but is instead exposed to the full diversity and richness of the target domain through the captions of unlabeled images. This enables the QA Generator to produce question-answer pairs that are more reflective of domain-specific semantics present in the unlabeled corpus. Hence, the resulting pseudo dataset $\hat{D}_u$ would be more accurate and reliable with our proposed \namestageaa, benefiting the subsequent learning of the VQA model in Equation~\ref{eq:vqa}. 

\section{Experiments}

\subsection{Datasets}

\paragraph{Kvasir-VQA}

Kvasir-VQA~\cite{gautam2024kvasir} is a large-scale visual question answering dataset curated for research in the gastrointestinal (GI) medical imaging domain. The dataset consists of 6,500 endoscopic images sourced from the HyperKvasir and Kvasir-Instrument datasets, covering a wide spectrum of clinically relevant GI findings, anatomical sites, and medical instruments. The images in Kvasir-VQA reflect real-world clinical scenarios, including both normal findings and a variety of pathological conditions such as polyps, ulcers, and esophagitis, as well as procedure-related scenes featuring different instruments and interventions. Each image is paired with one or more expert-annotated question-answer (QA) pairs. These QA annotations are diverse, encompassing multiple question types including yes/no, multiple-choice, and counting, thus enabling comprehensive evaluation of both recognition and reasoning abilities for VQA.

To standardize the evaluation process, we further convert all question-answer pairs in Kvasir-VQA into a multiple-choice format. For each question, we define the set of candidate options as all possible answers that appear for that question type throughout the dataset, so that the number of answer choices is not fixed for different questions. For experimental setup, we partition the dataset into training and testing splits, containing 18,499 and 18,075 QA pairs, respectively. To simulate the limited annotation scenario when learning our \nameshort~framework, only 1\% QA pairs in the training split are used as labeled data while the remaining training images are treated as unlabeled data.
\begin{table*}[t!]
    \centering
    \caption{Quantitative results of different learning strategies on the Kvasir-VQA dataset. All scores are reported in percentage (\%) of VQA accuracy. Only 1\% data are labeled during training and we use NVILA-Lite-2B as the MLLM backbone.}
    \vspace{-2mm}
	\resizebox{0.75\textwidth}{!}{
    
    \begin{tabular}{l c c c c c}
    
    \toprule 

    \multirow{2}{*}{Method} & \multicolumn{4}{c}{Image Category} & \multirow{2}{*}{Average}\\
    \cline{2-5}
    & Colitis & Esophagitis & Instrument & Polyps & \\
    \midrule
    Zero-Shot& 35.4 & 21.1 & 49.0 & 47.8 & 38.3 \\
    LoRA & 95.6 & 57.7 & 48.9 & 47.3 & 62.4 \\
    Full-Tuning & 89.0 & 60.8 & 51.2 & 51.3 & 63.1 \\
    \nameshort~(Ours) & 95.8 & 83.9 & 61.6 & 65.5 & 76.7 \\

    \midrule
    Zero-Shot (Qwen2.5-VL-72B) & 59.7 & 48.6 & 53.1 & 55.9 & 54.3 \\
    Fully-Supervised & 97.4 & 97.4 & 85.8 & 82.1 & 90.7 \\
    
    \bottomrule
    \label{tab:medical}
    \end{tabular}
    }
    \vspace{-3mm}
\end{table*}
\begin{table*}[t!]
    \centering
    \vspace{2mm}
    \caption{Quantitative results of different learning strategies on the SPORTU dataset. All scores are reported in percentage (\%) of VQA accuracy. ``Easy'', ``Medium'', and ``Hard'' denote different level of difficulty for sport understanding and reasoning questions. Only 1\% data are labeled during training and we use NVILA-Lite-2B as the MLLM backbone.}
    \vspace{-2mm}
	\resizebox{0.48\textwidth}{!}{
    
    \begin{tabular}{l c c c c}
    
    \toprule 

    Method & Easy & Medium & Hard & Average \\
    \midrule
    Zero-Shot & 50.5  & 34.0  & 37.5  & 40.7 \\
    LoRA & 82.5  & 59.2  & 22.3  & 54.7 \\
    Full-Tuning & 82.7  & 59.0  & 21.3  & 54.3 \\
    \nameshort~(Ours) & 82.5  & 60.4  & 46.3  & 63.1 \\

    \midrule
    Fully-Supervised & 98.1  & 75.4  & 66.1  & 79.9 \\
    
    \bottomrule 
    \vspace{-4mm}
    \label{tab:sport}
    \end{tabular}
    }
\end{table*}

\paragraph{SPORTU} SPORTU~\cite{xia2024sportu} is a recently released benchmark designed to evaluate the sports understanding and reasoning abilities of multimodal large language models (MLLMs). The dataset comprises 1,701 slow-motion sports video clips, spanning seven popular sports: \textit{American football, badminton, baseball, basketball, ice hockey, soccer}, and \textit{volleyball}. The questions in SPORTU are diverse and challenging, covering rule comprehension, tactical analysis, prediction of outcomes, and recognition of actions and fouls. All questions are categorized into three levels of difficulty: \textbf{easy}, which focuses on basic recognition tasks such as identifying the sport or counting players; \textbf{medium}, which requires knowledge of player roles and basic tactics; and \textbf{hard}, which involves deep reasoning about rules, foul detection, and scenario-based understanding. This tiered design enables comprehensive evaluation of MLLMs, from simple perception (in-distribution) to advanced, domain-specific sports reasoning (out-of-distribution). We partition the dataset into training and testing splits, containing 5,525 and 5,478 QA pairs, respectively. Similarly, we only consider 1\% QA pairs in the training split when learning our \nameshort~framework, while the remaining training images are treated as unlabeled data.

\subsection{Implementation Details}
Our entire implementation is based on the PyTorch framework. For simplicity, we use NVILA-Lite-2B~\cite{liu2025nvila} as our MLLM backbone for both our QA Generator and the VQA model. As for the large MLLM used for captioning distillation, we consider the state-of-the-art open source model, Qwen2.5-VL-72B-Instruct~\cite{bai2025qwen2}. All the model weights are initialized with official pretrained checkpoints. During training, we use the AdamW optimizer with an initial learning rate of $0.00001$ and a cosine annealing schedule for learning rate decay. The batch size is set as $16$. As for neuron selection, we set the number of parameter $K=1000$ on the Kvasir-VQA dataset and $K=3000$ for SPORTU, respectively. The training is performed on 16 NVIDIA A100 GPUs with 80GB memory each. During inference, we use deterministic (greedy) decoding for all answer generation for the VQA model, i.e., at each generation step, the model always selects the token with the highest probability. This ensures that the outputs are fully reproducible and comparable across runs. As for pseudo QA Generation, we choose to use nucleus sampling on our QA Generator to produce several different question-answer pairs for each single visual input.

\subsection{Quantitative and Qualitative Experiments}

While multimodal large language models (MLLMs) have demonstrated remarkable abilities on general visual understanding benchmarks, their robustness and adaptability often fall short when transferred to specialized domains such as medical image question answering. This gap arises primarily from two factors: the limited availability of expert-annotated data, and the significant distribution shift between general pretraining data and domain-specific imagery. In the context of medical imaging, acquiring comprehensive labeled datasets is both costly and time-consuming, which motivates the need for approaches that can effectively leverage abundant unlabeled data. To evaluate our method under such kind of scenarios, we design experiments on the Kvasir-VQA dataset, a large-scale benchmark of expert-annotated QA pairs paired with real-world endoscopic images.

\begin{table*}[t!]
    \centering
    \caption{Ablation study of our \namestageaa~for the QA Generator on the Kvasir-VQA dataset.}
    \vspace{-2mm}
	\resizebox{0.8\textwidth}{!}{
    
    \begin{tabular}{l c c c c c}
    
    \toprule 

    \multirow{2}{*}{Method} & \multicolumn{4}{c}{Image Category} & \multirow{2}{*}{Average}\\
    \cline{2-5}
    & Colitis & Esophagitis & Instrument & Polyps & \\
    \midrule

    Baseline & 93.0 & 88.2 & 55.7 & 56.2 & 73.3 \\
    Baseline+Distill. & 95.1 & 80.3 & 59.7 & 60.4 & 73.9 \\
    Baseline+Distill.+QA Neurons & 95.8 & 83.9 & 61.6 & 65.5 & 76.7 \\
    
    \bottomrule 
    \label{tab:ab1}
    \end{tabular}
    }
    \vspace{-5mm}
\end{table*}
\begin{table*}[t!]
    \centering
    \vspace{2mm}
    \caption{Ablation study of our \namestageaa~for the QA Generator on the SPORTU dataset.}
    \vspace{-2mm}
	\resizebox{0.63\textwidth}{!}{
    
    \begin{tabular}{l c c c c}
    
    \toprule 

    Method & Easy & Medium & Hard & Average \\
    \midrule
    Baseline & 75.5 & 60.3 & 44.4 & 60.1 \\
    Baseline+Distill.& 82.9 & 61.2 & 42.3 & 62.2 \\
    Baseline+Distill.+QA Neurons  & 82.5  & 60.4  & 46.3  & 63.1 \\
    
    \bottomrule 
    \label{tab:ab2}
    \end{tabular}
    }
    \vspace{-6mm}
\end{table*}

Table~\ref{tab:medical} presents a detailed comparison of different learning strategies on the Kvasir-VQA datasets, where only $1\%$ of the training data is labeled and the remaining ones are treated as unlabeled. From this Table, we see that zero-shot inference using NVILA-Lite-2B achieves only $38.3\%$ average accuracy, indicating the difficulty of the task without any fine-tuning. Fine-tuning with LoRA~\cite{hu2022lora} or full parameter updates on just $1\%$ labeled data yields moderate improvements ($62.4\%$ and $63.1\%$ accuracy, respectively), but these methods still struggle on less-represented or out-of-distribution categories. By augmenting the training with pseudo QA pairs generated from the unlabeled images, our approach significantly boosts average accuracy to $76.7\%$, especially on challenging categories such as Esophagitis. As a reference, the fully supervised model (using all labeled data) achieves $90.7\%$.

\begin{figure}[t]
  \centering
  \includegraphics[width=\linewidth]{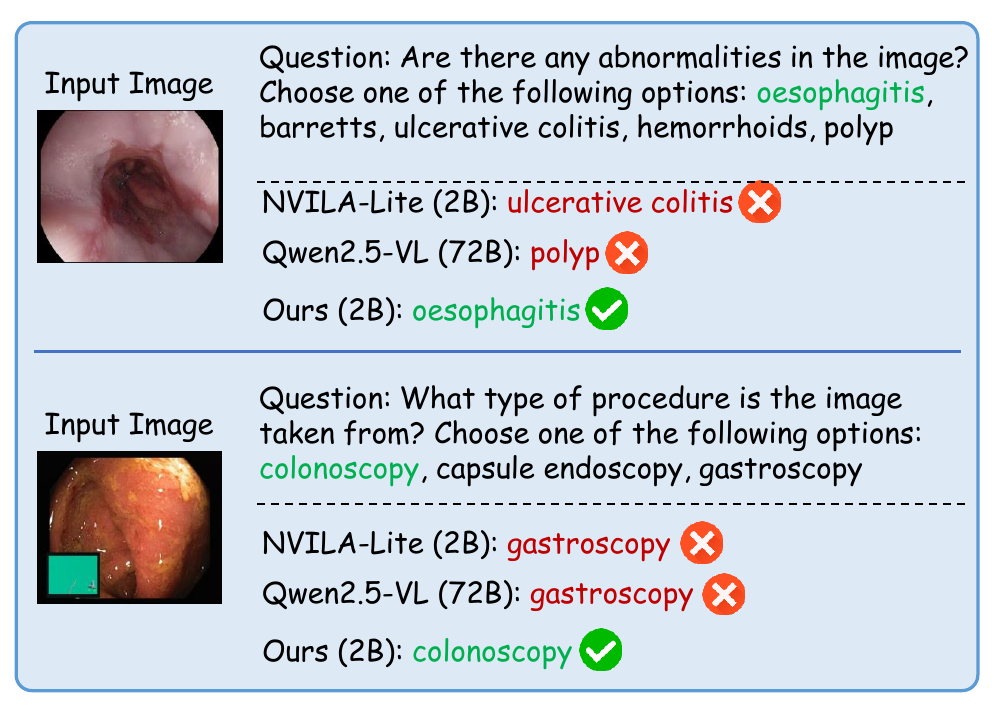}
  \vspace{-6mm}\caption{Qualitative results on the Kvasir-VQA dataset.}
  \label{figure:medicalg}
  \vspace{-4mm}
\end{figure}

\begin{figure}[!t]
  \centering
  \includegraphics[width=\linewidth]{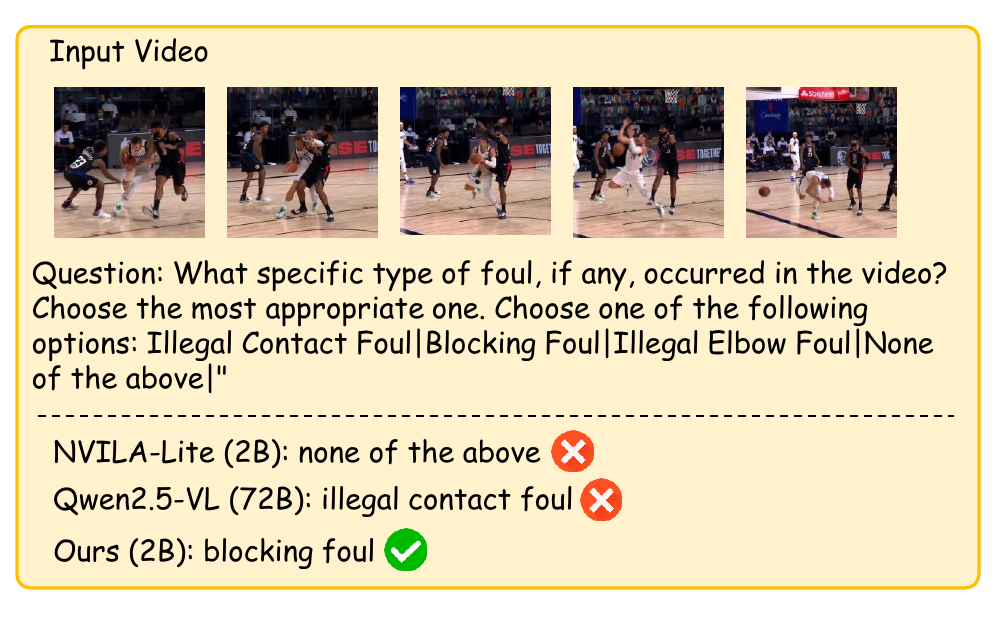}
  \vspace{-6mm}\caption{Qualitative results on the SPORTU dataset.}
  \label{figure:sport}
  \vspace{-4mm}
\end{figure}

We further evaluate our method on the SPORTU dataset to assess its generalization ability in the sports domain, which poses unique challenges compared to the medical setting. Unlike static medical images, SPORTU consists of dynamic sports video clips that require not only visual recognition but also temporal reasoning, action understanding, and comprehension of complex game rules. These factors make sports VQA a particularly demanding task for multimodal large language models. As shown in Table~\ref{tab:sport}, our \nameshort~framework achieves consistent improvements over all baselines. In particular, our method raises the average accuracy to 63.1\%, outperforming both LoRA and full-tuning by a clear margin. Notably, the accuracy on “Hard” questions—which require advanced reasoning—improves significantly from around 22.3\% to 46.3\%. In addition to quantitative results, we also provide qualitative comparisons as shown in Figure~\ref{figure:medical} and~\ref{figure:sport}. We see that, our \nameshort~framework is able to produce accurate answers compared to state-of-the-art MLLMs on the challenging medical and sport domains. Through the above experiments, we verify that our ~\nameshort~framework, which directly incorporates unlabeled data through pseudo QA generation, can substantially enhance VQA performance on out-of-distribution domain with limited annotations.

\subsection{Ablation Studies}
As shown in Table~\ref{tab:ab1}, our ablation study on Kvasir-VQA evaluates the effects of caption distillation and selective neuron updates. Caption distillation alone leads to only a minor increase in average accuracy (an improvement of just 0.6\% over the baseline). In contrast, when selective neuron updates are applied, the accuracy increases substantially, yielding a total gain of 3.4\% over the baseline. A similar trend is observed on the SPORTU dataset (Table~\ref{tab:ab2}). These results highlight that targeted neuron selection is crucial for effective distillation for the QA Generator, and that substantial performance gains can only be achieved when both components are used together.

\section{Conclusion}
In this paper, we introduce \nameshort, a label-efficient adaptation framework designed to transfer MLLMs to out-of-distribution (OOD) domains that extend far beyond their original pretraining distribution. Our approach first utilizes \namestagea~to generate domain-relevant pseudo question-answer pairs from large pools of unlabeled visual data, effectively expanding the training set in scenarios where annotated examples are extremely limited. In addition, we employ \namestageaa~for the QA Generator, a selective neuron updating strategy that identifies and updates only knowledge-relevant neurons while acquiring domain knowledge from state-of-the-art large MLLMs. We conduct extensive experiments on OOD benchmarks, including gastrointestinal endoscopy and sports VQA, demonstrating the applicability of our framework. Detailed ablation studies and both quantitative and qualitative analyses consistently show that our proposed \nameshort~framework achieves substantial improvements over conventional fine-tuning methods, validating its effectiveness and robustness for adapting MLLMs to challenging domain-specific visual tasks with limited annotations.

\bibliography{aaai2026}

\begin{thebibliography}{36}
\providecommand{\natexlab}[1]{#1}

\bibitem[{Achiam et~al.(2023)Achiam, Adler, Agarwal, Ahmad, Akkaya, Aleman, Almeida, Altenschmidt, Altman, Anadkat et~al.}]{achiam2023gpt}
Achiam, J.; Adler, S.; Agarwal, S.; Ahmad, L.; Akkaya, I.; Aleman, F.~L.; Almeida, D.; Altenschmidt, J.; Altman, S.; Anadkat, S.; et~al. 2023.
\newblock Gpt-4 technical report.
\newblock \emph{arXiv preprint arXiv:2303.08774}.

\bibitem[{Alayrac et~al.(2022)Alayrac, Donahue, Luc, Miech, Barr, Hasson, Lenc, Mensch, Millican, Reynolds et~al.}]{alayrac2022flamingo}
Alayrac, J.-B.; Donahue, J.; Luc, P.; Miech, A.; Barr, I.; Hasson, Y.; Lenc, K.; Mensch, A.; Millican, K.; Reynolds, M.; et~al. 2022.
\newblock Flamingo: a visual language model for few-shot learning.
\newblock \emph{Advances in neural information processing systems}, 35: 23716--23736.

\bibitem[{Bai et~al.(2023)Bai, Bai, Chu, Cui, Dang, Deng, Fan, Ge, Han, Huang et~al.}]{bai2023qwen}
Bai, J.; Bai, S.; Chu, Y.; Cui, Z.; Dang, K.; Deng, X.; Fan, Y.; Ge, W.; Han, Y.; Huang, F.; et~al. 2023.
\newblock Qwen technical report.
\newblock \emph{arXiv preprint arXiv:2309.16609}.

\bibitem[{Bai et~al.(2025)Bai, Chen, Liu, Wang, Ge, Song, Dang, Wang, Wang, Tang et~al.}]{bai2025qwen2}
Bai, S.; Chen, K.; Liu, X.; Wang, J.; Ge, W.; Song, S.; Dang, K.; Wang, P.; Wang, S.; Tang, J.; et~al. 2025.
\newblock Qwen2. 5-vl technical report.
\newblock \emph{arXiv preprint arXiv:2502.13923}.

\bibitem[{Bhatia et~al.(2024)Bhatia, Nagoudi, Cavusoglu, and Abdul-Mageed}]{bhatia2024fintral}
Bhatia, G.; Nagoudi, E. M.~B.; Cavusoglu, H.; and Abdul-Mageed, M. 2024.
\newblock Fintral: A family of gpt-4 level multimodal financial large language models.
\newblock \emph{arXiv preprint arXiv:2402.10986}.

\bibitem[{Cheng, Huang, and Wei(2023)}]{cheng2023adapting}
Cheng, D.; Huang, S.; and Wei, F. 2023.
\newblock Adapting large language models via reading comprehension.
\newblock In \emph{The Twelfth International Conference on Learning Representations}.

\bibitem[{Cheng et~al.(2024)Cheng, Huang, Zhu, Zhang, Zhao, Luan, Dai, and Zhang}]{cheng2024domain}
Cheng, D.; Huang, S.; Zhu, Z.; Zhang, X.; Zhao, W.~X.; Luan, Z.; Dai, B.; and Zhang, Z. 2024.
\newblock On Domain-Adaptive Post-Training for Multimodal Large Language Models.
\newblock \emph{arXiv preprint arXiv:2411.19930}.

\bibitem[{Dai et~al.(2021)Dai, Dong, Hao, Sui, Chang, and Wei}]{dai2021knowledge}
Dai, D.; Dong, L.; Hao, Y.; Sui, Z.; Chang, B.; and Wei, F. 2021.
\newblock Knowledge neurons in pretrained transformers.
\newblock \emph{arXiv preprint arXiv:2104.08696}.

\bibitem[{Fang et~al.(2024)Fang, Bi, Wang, Jiang, Gao, Wang, Zhang, Shi, Wang, and Chua}]{fang2024towards}
Fang, J.; Bi, Z.; Wang, R.; Jiang, H.; Gao, Y.; Wang, K.; Zhang, A.; Shi, J.; Wang, X.; and Chua, T.-S. 2024.
\newblock Towards neuron attributions in multi-modal large language models.
\newblock \emph{Advances in Neural Information Processing Systems}, 37: 122867--122890.

\bibitem[{Gautam et~al.(2024)Gautam, Stor{\aa}s, Midoglu, Hicks, Thambawita, Halvorsen, and Riegler}]{gautam2024kvasir}
Gautam, S.; Stor{\aa}s, A.~M.; Midoglu, C.; Hicks, S.~A.; Thambawita, V.; Halvorsen, P.; and Riegler, M.~A. 2024.
\newblock Kvasir-vqa: A text-image pair gi tract dataset.
\newblock In \emph{Proceedings of the First International Workshop on Vision-Language Models for Biomedical Applications}, 3--12.

\bibitem[{Geva et~al.(2022)Geva, Caciularu, Wang, and Goldberg}]{geva2022transformer}
Geva, M.; Caciularu, A.; Wang, K.~R.; and Goldberg, Y. 2022.
\newblock Transformer feed-forward layers build predictions by promoting concepts in the vocabulary space.
\newblock \emph{arXiv preprint arXiv:2203.14680}.

\bibitem[{Geva et~al.(2020)Geva, Schuster, Berant, and Levy}]{geva2020transformer}
Geva, M.; Schuster, R.; Berant, J.; and Levy, O. 2020.
\newblock Transformer feed-forward layers are key-value memories.
\newblock \emph{arXiv preprint arXiv:2012.14913}.

\bibitem[{Hu et~al.(2022)Hu, Shen, Wallis, Allen-Zhu, Li, Wang, Wang, Chen et~al.}]{hu2022lora}
Hu, E.~J.; Shen, Y.; Wallis, P.; Allen-Zhu, Z.; Li, Y.; Wang, S.; Wang, L.; Chen, W.; et~al. 2022.
\newblock Lora: Low-rank adaptation of large language models.
\newblock \emph{ICLR}, 1(2): 3.

\bibitem[{Kaplan et~al.(2020)Kaplan, McCandlish, Henighan, Brown, Chess, Child, Gray, Radford, Wu, and Amodei}]{kaplan2020scaling}
Kaplan, J.; McCandlish, S.; Henighan, T.; Brown, T.~B.; Chess, B.; Child, R.; Gray, S.; Radford, A.; Wu, J.; and Amodei, D. 2020.
\newblock Scaling laws for neural language models.
\newblock \emph{arXiv preprint arXiv:2001.08361}.

\bibitem[{Kim et~al.(2025)Kim, Park, Lee, Yeo, and Hwang}]{kim2025videoicl}
Kim, K.; Park, G.; Lee, Y.; Yeo, W.; and Hwang, S.~J. 2025.
\newblock VideoICL: Confidence-based Iterative In-context Learning for Out-of-Distribution Video Understanding.
\newblock In \emph{Proceedings of the Computer Vision and Pattern Recognition Conference}, 3295--3305.

\bibitem[{Lee et~al.(2013)}]{lee2013pseudo}
Lee, D.-H.; et~al. 2013.
\newblock Pseudo-label: The simple and efficient semi-supervised learning method for deep neural networks.
\newblock In \emph{Workshop on challenges in representation learning, ICML}, volume~3, 896. Atlanta.

\bibitem[{Lei et~al.(2023)Lei, Li, Shen, Zhang, and Shan}]{lei2023clip}
Lei, Y.; Li, Z.; Shen, Y.; Zhang, J.; and Shan, H. 2023.
\newblock Clip-lung: Textual knowledge-guided lung nodule malignancy prediction.
\newblock In \emph{International Conference on Medical Image Computing and Computer-Assisted Intervention}, 403--412. Springer.

\bibitem[{Liu et~al.(2023{\natexlab{a}})Liu, Li, Wu, and Lee}]{liu2023visual}
Liu, H.; Li, C.; Wu, Q.; and Lee, Y.~J. 2023{\natexlab{a}}.
\newblock Visual instruction tuning.
\newblock \emph{Advances in neural information processing systems}, 36: 34892--34916.

\bibitem[{Liu et~al.(2023{\natexlab{b}})Liu, Zhang, Chen, Xiao, Lu, A~Landman, Yuan, Yuille, Tang, and Zhou}]{liu2023clip}
Liu, J.; Zhang, Y.; Chen, J.-N.; Xiao, J.; Lu, Y.; A~Landman, B.; Yuan, Y.; Yuille, A.; Tang, Y.; and Zhou, Z. 2023{\natexlab{b}}.
\newblock Clip-driven universal model for organ segmentation and tumor detection.
\newblock In \emph{Proceedings of the IEEE/CVF international conference on computer vision}, 21152--21164.

\bibitem[{Liu et~al.(2025)Liu, Zhu, Shi, Zhang, Lou, Yang, Xi, Cao, Gu, Li et~al.}]{liu2025nvila}
Liu, Z.; Zhu, L.; Shi, B.; Zhang, Z.; Lou, Y.; Yang, S.; Xi, H.; Cao, S.; Gu, Y.; Li, D.; et~al. 2025.
\newblock Nvila: Efficient frontier visual language models.
\newblock In \emph{Proceedings of the Computer Vision and Pattern Recognition Conference}, 4122--4134.

\bibitem[{Ming et~al.(2022)Ming, Cai, Gu, Sun, Li, and Li}]{ming2022delving}
Ming, Y.; Cai, Z.; Gu, J.; Sun, Y.; Li, W.; and Li, Y. 2022.
\newblock Delving into out-of-distribution detection with vision-language representations.
\newblock \emph{Advances in neural information processing systems}, 35: 35087--35102.

\bibitem[{Pan et~al.(2023)Pan, Cao, Wang, Yang, and Wang}]{pan2023finding}
Pan, H.; Cao, Y.; Wang, X.; Yang, X.; and Wang, M. 2023.
\newblock Finding and editing multi-modal neurons in pre-trained transformers.
\newblock \emph{arXiv preprint arXiv:2311.07470}.

\bibitem[{Sun et~al.(2023)Sun, Liu, Bair, and Kolter}]{sun2023simple}
Sun, M.; Liu, Z.; Bair, A.; and Kolter, J.~Z. 2023.
\newblock A simple and effective pruning approach for large language models.
\newblock \emph{arXiv preprint arXiv:2306.11695}.

\bibitem[{Touvron et~al.(2023)Touvron, Lavril, Izacard, Martinet, Lachaux, Lacroix, Rozi{\`e}re, Goyal, Hambro, Azhar et~al.}]{touvron2023llama}
Touvron, H.; Lavril, T.; Izacard, G.; Martinet, X.; Lachaux, M.-A.; Lacroix, T.; Rozi{\`e}re, B.; Goyal, N.; Hambro, E.; Azhar, F.; et~al. 2023.
\newblock Llama: Open and efficient foundation language models.
\newblock \emph{arXiv preprint arXiv:2302.13971}.

\bibitem[{Wang et~al.(2023)Wang, Li, Yao, and Li}]{wang2023clipn}
Wang, H.; Li, Y.; Yao, H.; and Li, X. 2023.
\newblock Clipn for zero-shot ood detection: Teaching clip to say no.
\newblock In \emph{Proceedings of the IEEE/CVF International Conference on Computer Vision}, 1802--1812.

\bibitem[{Wang et~al.(2024{\natexlab{a}})Wang, Bai, Tan, Wang, Fan, Bai, Chen, Liu, Wang, Ge et~al.}]{wang2024qwen2}
Wang, P.; Bai, S.; Tan, S.; Wang, S.; Fan, Z.; Bai, J.; Chen, K.; Liu, X.; Wang, J.; Ge, W.; et~al. 2024{\natexlab{a}}.
\newblock Qwen2-vl: Enhancing vision-language model's perception of the world at any resolution.
\newblock \emph{arXiv preprint arXiv:2409.12191}.

\bibitem[{Wang et~al.(2024{\natexlab{b}})Wang, Chen, Han, Lin, Zhao, Liu, Zhai, Yuan, You, and Yang}]{wang2024exploring}
Wang, Y.; Chen, W.; Han, X.; Lin, X.; Zhao, H.; Liu, Y.; Zhai, B.; Yuan, J.; You, Q.; and Yang, H. 2024{\natexlab{b}}.
\newblock Exploring the reasoning abilities of multimodal large language models (mllms): A comprehensive survey on emerging trends in multimodal reasoning.
\newblock \emph{arXiv preprint arXiv:2401.06805}.

\bibitem[{Xia et~al.(2024)Xia, Yang, Zou, Tracy, Wang, Lu, Lai, He, Shao, Xie et~al.}]{xia2024sportu}
Xia, H.; Yang, Z.; Zou, J.; Tracy, R.; Wang, Y.; Lu, C.; Lai, C.; He, Y.; Shao, X.; Xie, Z.; et~al. 2024.
\newblock Sportu: A comprehensive sports understanding benchmark for multimodal large language models.
\newblock \emph{arXiv preprint arXiv:2410.08474}.

\bibitem[{Xie et~al.(2020)Xie, Luong, Hovy, and Le}]{xie2020self}
Xie, Q.; Luong, M.-T.; Hovy, E.; and Le, Q.~V. 2020.
\newblock Self-training with noisy student improves imagenet classification.
\newblock In \emph{Proceedings of the IEEE/CVF conference on computer vision and pattern recognition}, 10687--10698.

\bibitem[{Yang et~al.(2023)Yang, Qi, Feng, Zhang, and Shi}]{yang2023revisiting}
Yang, L.; Qi, L.; Feng, L.; Zhang, W.; and Shi, Y. 2023.
\newblock Revisiting weak-to-strong consistency in semi-supervised semantic segmentation.
\newblock In \emph{Proceedings of the IEEE/CVF conference on computer vision and pattern recognition}, 7236--7246.

\bibitem[{Yu and Ananiadou(2023)}]{yu2023neuron}
Yu, Z.; and Ananiadou, S. 2023.
\newblock Neuron-level knowledge attribution in large language models.
\newblock \emph{arXiv preprint arXiv:2312.12141}.

\bibitem[{Yu and Ananiadou(2025)}]{yu2025locate}
Yu, Z.; and Ananiadou, S. 2025.
\newblock Locate-then-Merge: Neuron-Level Parameter Fusion for Mitigating Catastrophic Forgetting in Multimodal LLMs.
\newblock \emph{arXiv preprint arXiv:2505.16703}.

\bibitem[{Zeng et~al.(2023)Zeng, Xie, Lu, and Xia}]{zeng2023pefat}
Zeng, Q.; Xie, Y.; Lu, Z.; and Xia, Y. 2023.
\newblock Pefat: Boosting semi-supervised medical image classification via pseudo-loss estimation and feature adversarial training.
\newblock In \emph{Proceedings of the IEEE/CVF conference on computer vision and pattern recognition}, 15671--15680.

\bibitem[{Zhang et~al.(2023{\natexlab{a}})Zhang, Xu, Usuyama, Xu, Bagga, Tinn, Preston, Rao, Wei, Valluri et~al.}]{zhang2023biomedclip}
Zhang, S.; Xu, Y.; Usuyama, N.; Xu, H.; Bagga, J.; Tinn, R.; Preston, S.; Rao, R.; Wei, M.; Valluri, N.; et~al. 2023{\natexlab{a}}.
\newblock Biomedclip: a multimodal biomedical foundation model pretrained from fifteen million scientific image-text pairs.
\newblock \emph{arXiv preprint arXiv:2303.00915}.

\bibitem[{Zhang et~al.(2023{\natexlab{b}})Zhang, Wang, Cheng, Kurohashi et~al.}]{zhang2023reformulating}
Zhang, Y.; Wang, Y.; Cheng, F.; Kurohashi, S.; et~al. 2023{\natexlab{b}}.
\newblock Reformulating domain adaptation of large language models as adapt-retrieve-revise: A case study on Chinese legal domain.
\newblock \emph{arXiv preprint arXiv:2310.03328}.

\bibitem[{Zhang et~al.(2024)Zhang, Zhang, Gao, Zhang, Shutova, Zhou, and Zhang}]{zhang2024gradient}
Zhang, Z.; Zhang, Q.; Gao, Z.; Zhang, R.; Shutova, E.; Zhou, S.; and Zhang, S. 2024.
\newblock Gradient-based parameter selection for efficient fine-tuning.
\newblock In \emph{Proceedings of the IEEE/CVF Conference on Computer Vision and Pattern Recognition}, 28566--28577.

\end{thebibliography}
\clearpage
\appendix
\section{Additional Visualization}

\subsection{Comparison with Baseline in VQA}
\begin{figure}[t]
  \centering
  \includegraphics[width=\linewidth]{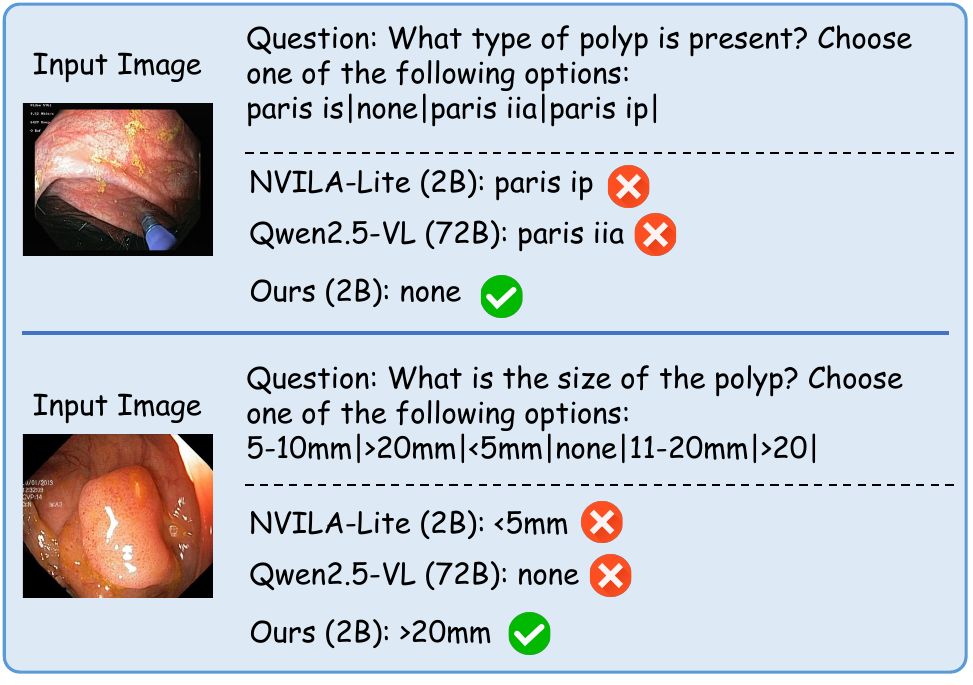}
  \vspace{-6mm}\caption{Qualitative results on the Kvasir-VQA dataset.}
  \vspace{-4mm}
\end{figure}
\begin{figure}[!t]
  \centering
  \includegraphics[width=\linewidth]{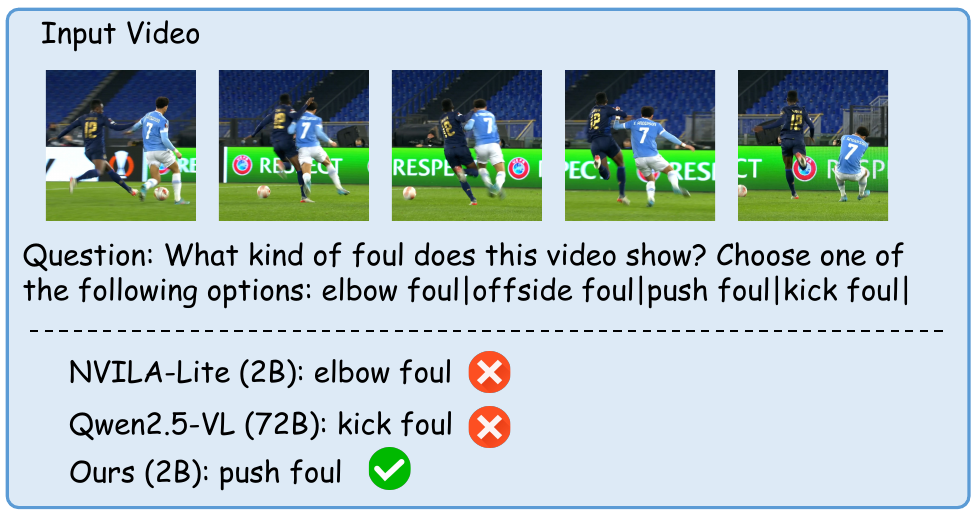}
  \vspace{-6mm}\caption{Qualitative results on the SPORTU dataset.}
\end{figure}

We provide additional qualitative results comparing our LEAML method with baseline approaches on both medical and sports VQA tasks. These examples demonstrate our method's superior performance in generating accurate, domain-specific answers. The results show that our approach better understands specialized terminology and visual patterns, leading to more precise responses compared to standard MLLMs.

\subsection{Generated Pseudo-QA Examples}
\begin{figure}[t]
  \centering
  \includegraphics[width=\linewidth]{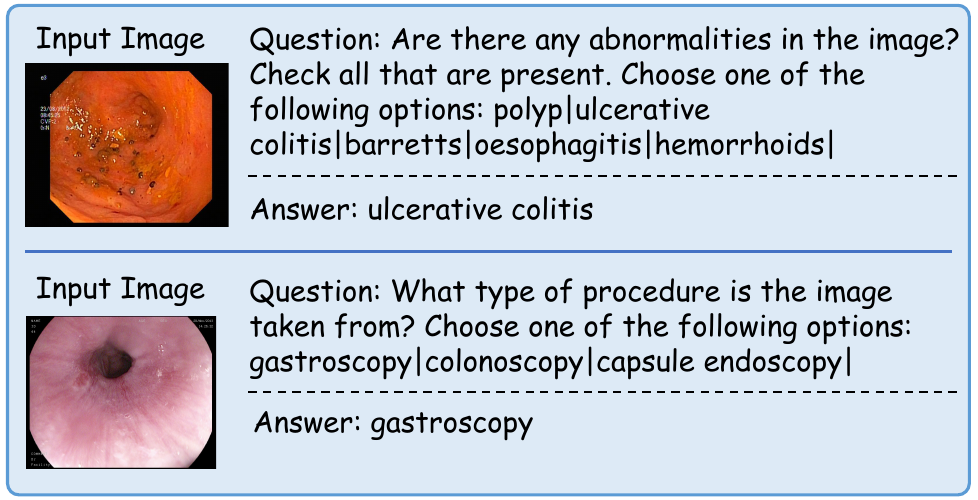}
  \vspace{-6mm}\caption{Qualitative results of generated pseudo-QA on the Kvasir-VQA dataset.}
  \label{figure:medicalb}
\end{figure}
\begin{figure}[t]
  \centering
  \includegraphics[width=\linewidth]{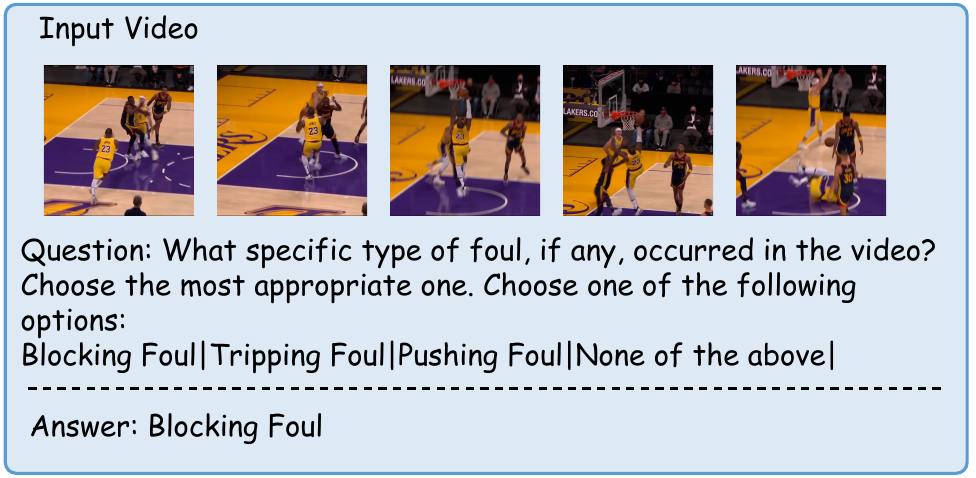}
  \vspace{-6mm}\caption{Qualitative results of generated pseudo-QA on the SPORTU dataset.}
  \label{figure:medicalc}
\end{figure}
\begin{figure}[!ht]
  \centering
  \includegraphics[width=\linewidth]{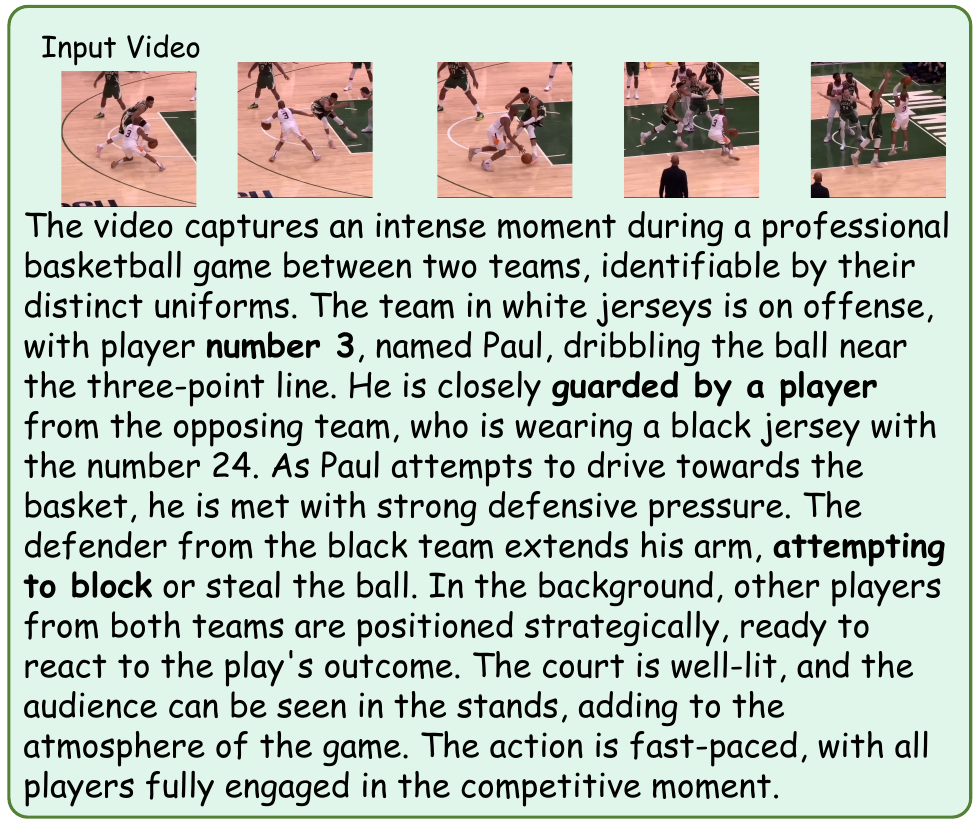}
  \vspace{-6mm}\caption{Qualitative results of generated caption on the SPORTU dataset.}
  \label{figure:medical}
  \vspace{-4mm}
\end{figure}

We show examples of pseudo question-answer pairs generated by our QA Generator. These synthetic QA pairs effectively expand the training dataset and provide domain-relevant supervision for the subsequent fine-tuning stage. The generated questions are contextually appropriate and cover diverse aspects of the visual content, while the answers demonstrate proper domain knowledge and terminology usage.

\subsection{Captions from Large MLLM}

\begin{figure}[t]
  \centering
  \includegraphics[width=\linewidth]{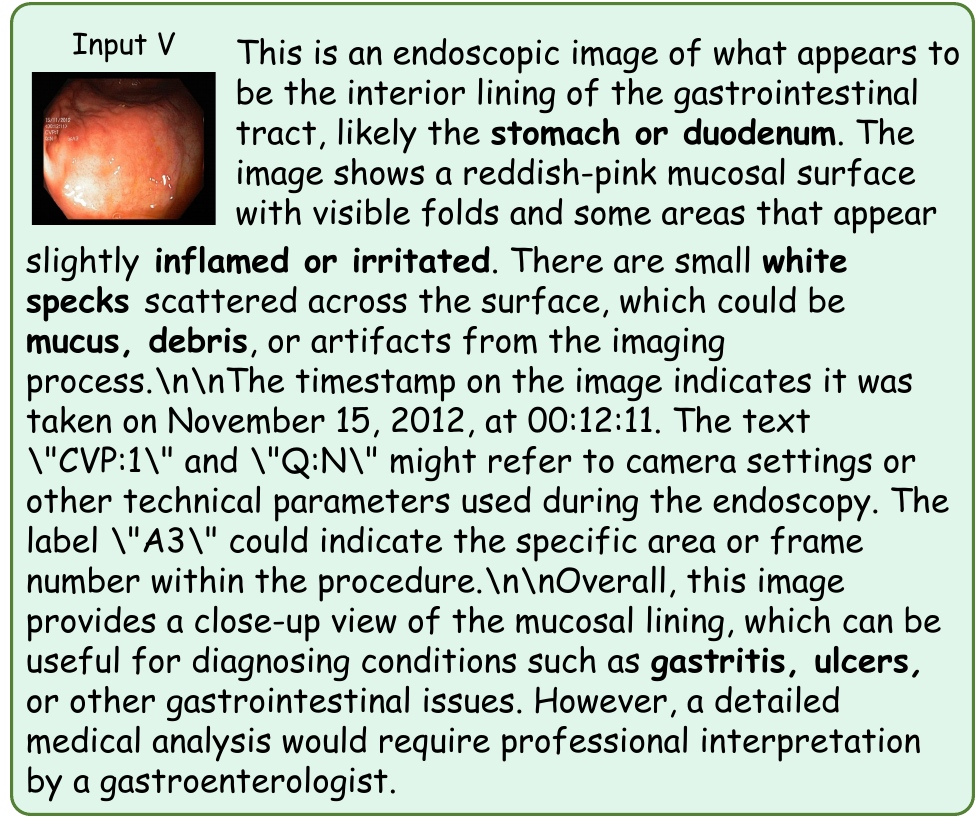}
  \vspace{-6mm}\caption{Qualitative results of generated caption on the Kvasir-VQA dataset.}
  \label{figure:medicala}
  \vspace{-4mm}
\end{figure}

We present captions generated by the large MLLM (Qwen2.5-VL-72B) used in our distillation process. These captions successfully capture important visual details and domain-specific information, which helps improve the QA Generator's learning through knowledge distillation.

\begin{figure}[t]
  \centering
  \includegraphics[width=\linewidth]{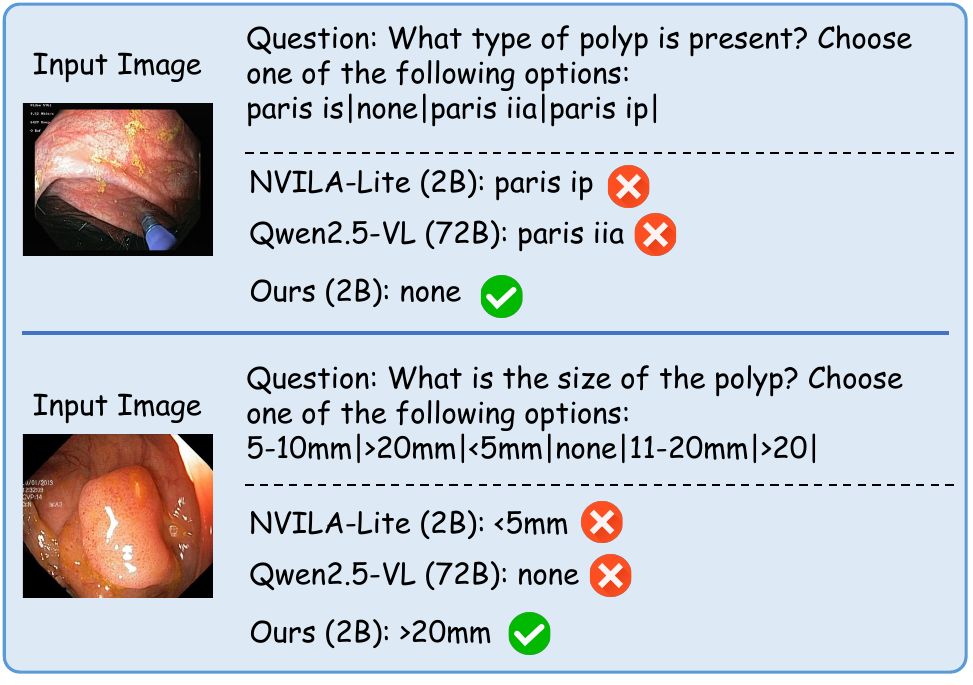}
  \vspace{-6mm}\caption{Qualitative results on the Kvasir-VQA dataset.}
  \vspace{-4mm}
\end{figure}
\begin{figure}[!t]
  \centering
  \includegraphics[width=\linewidth]{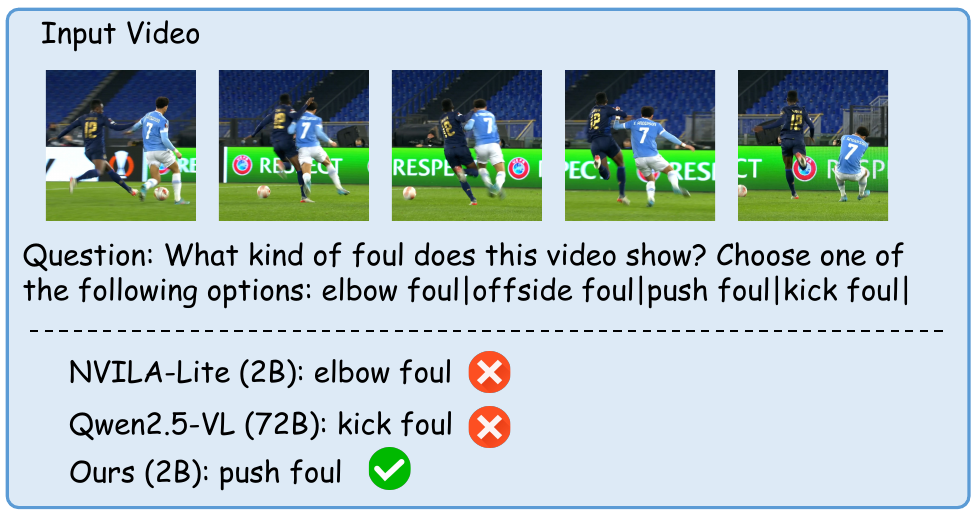}
  \vspace{-6mm}\caption{Qualitative results on the SPORTU dataset.}
\end{figure}
\end{document}